\documentclass[conference]{IEEEtran}
\IEEEoverridecommandlockouts

\usepackage{graphicx}
\usepackage{cite}
\usepackage{amsmath,amssymb}
\usepackage{hyperref}
\usepackage{xcolor} 
\usepackage{comment}
\usepackage{algorithm}
\usepackage{algorithmic}

\usepackage{graphicx}  
\usepackage{multirow}  
\usepackage{tcolorbox} 
\usepackage{array}     
\usepackage[normalem]{ulem}
\usepackage{subcaption}
\usepackage[inline]{enumitem}

\newcommand{\removed}[1]{{\color{gray} }}

\title{CompliantVLA-adaptor: VLM-Guided Variable Impedance Action for Safe Contact-Rich Manipulation}

\author{
    
    \IEEEauthorblockN{Heng Zhang*\textsuperscript{1,2,3}, Wei-Hsing Huang*\textsuperscript{4}, Qiyi Tong\textsuperscript{1,2}, Gokhan Solak\textsuperscript{1}, Puze Liu \textsuperscript{5}, Kaidi Zhang \textsuperscript{3},Sheng Liu\textsuperscript{7}, \\Jan Peters \textsuperscript{5,6}, Yu She\textsuperscript{3}, Arash Ajoudani\textsuperscript{1}}
     \IEEEauthorblockA{
         \textsuperscript{1}Human-Robot Interfaces and Interaction Lab, Istituto Italiano di Tecnologia, Genoa, Italy.\\
         \textsuperscript{2}Ph.D. program of national interest in Robotics and Intelligent Machines (DRIM) and Università di Genova, Genoa, Italy.\\
         \textsuperscript{3}Edwardson School of Industrial Engineering, Purdue University, West Lafayette, IN 47907, USA\\
         \textsuperscript{4}Georgia Institute of Technology, Atlanta, USA. \\
         \textsuperscript{5}German Research Center for AI, Germany \\
         \textsuperscript{6}TU Darmstadt, Darmstadt, Germany\\
         \textsuperscript{7}Karlsruhe Institute of Technology, Karlsruhe, Germany. \\
         *These two authors contributed equally to this work \\
     }
}

\begin{document}

\maketitle

\begin{abstract}
We propose a CompliantVLA-adaptor that augments the state-of-the-art Vision-Language-Action (VLA) models with vision-language model (VLM)-informed context-aware variable impedance control (VIC) to improve the safety and effectiveness of contact-rich robotic manipulation tasks. 
Existing VLA systems (e.g., RDT, Pi0.5, OpenVLA-oft) typically output position, but lack force-aware adaptation, leading to unsafe or failed interactions in physical tasks involving contact, compliance, or uncertainty. 
In the proposed CompliantVLA-adaptor, a VLM interprets task context from images and natural language to adapt the stiffness and damping parameters of a VIC controller. 
These parameters are further regulated using real-time force/torque feedback to ensure interaction forces remain within safe thresholds.
We demonstrate that our method outperforms the VLA baselines on a suite of complex contact-rich tasks, both in simulation and the real world, with improved success rates and reduced force violations.  
This work presents a promising path towards a safe foundation model for physical contact-rich manipulation.
We release our code, prompts, and force-torque-impedance-scenario context datasets at 
\url{https://sites.google.com/view/compliantvla}.
\end{abstract}


\section{Introduction}
Recent advances in Vision-Language-Action (VLA) models have enabled robots to understand and execute complex tasks described in natural language, such as
RDT~\cite{liu2024rdt}, Pi0\cite{black2024pi_0}, DiffusionVLA~\cite{wen2025diffusionvla}, Pi0.5~\cite{intelligence2025pi_}, OpenVLA~\cite{kim2024openvla} and OpenVLA-oft~\cite{kim2025fine}. 
These models leverage large-scale pretraining on diverse datasets to learn rich visual and linguistic representations, allowing them to generalize across a wide range of manipulation tasks. 

Current state-of-the-art VLA models demonstrate remarkable generalization across diverse manipulation tasks. However, these models fundamentally operate through position or trajectory control, treating the robot as a rigid position-tracking system that lacks consideration of the physical interaction dynamics involved in contact-rich tasks~\cite{zhong2025survey}.
We observe the limitation of the current VLA models that hinder their performance in contact-rich tasks if considering the contact force threshold (e.g., \textless 30N).
The contact force can become huge when the robot is executing raw output of VLA in contact-rich tasks without force regulation. Safety issue highlights the limitation of current VLA models~\cite{ma2024survey}.
\begin{figure}
    \centering
    \includegraphics[trim=0.1cm 0.1cm 0.1cm 0.1cm, width=0.9\linewidth]{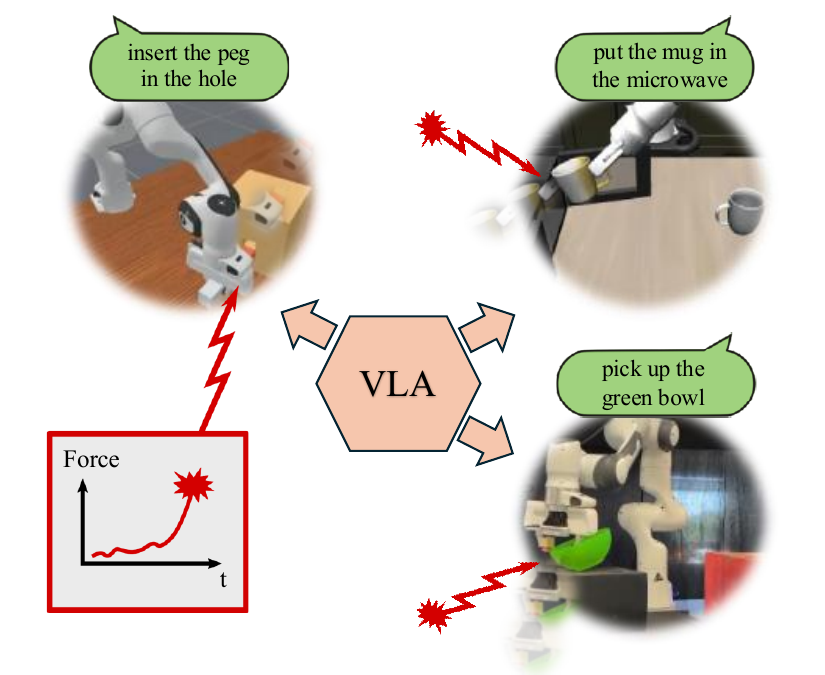}
    \caption{\small Existing VLA systems lack force-awareness, leading to unsafe interactions in physical tasks involving contact or uncertainty. We see this challenge as a promising direction for safer deployment of VLA systems.}
    \label{fig:motivation}
\end{figure}
This rigid execution paradigm leads to critical failures when interacting physically. 
They have a core limitation as illustrated in Fig.~\ref{fig:motivation}: VLAs excel at semantic understanding but cannot translate this understanding into the physical compliance required for safe execution. 
Consider a robot assembling a delicate electronic device (``insert the USB cable"): it must insert a fragile connector with just enough force to engage the latch, yet gentle enough to avoid damage. 
Or imagine a robot closing a drawer, where excessive force could damage the drawer or its contents, while insufficient force might leave the drawer ajar. 
The contact-rich manipulation tasks
require not just visual understanding, but also precise force regulation and adaptive compliance. 
While recent VLA models have revolutionized robot learning through large-scale pretraining~\cite{o2024open}, they critically lack the force-aware adaptation necessary for safe physical interaction.
Zhang et al.\cite{zhang2025ta} also point out the importance of torque-awareness in VLA models, but they only consider the torque prediction without force feedback and impedance control.

These shortcomings of the current VLA models lead
to position commands even when encountering unexpected resistance, potentially causing irreversible damage to delicate objects or the robot itself. The disconnect between high-level semantic comprehension and low-level force-aware control represents a critical barrier to deploying VLA systems in real-world scenarios where safe physical interaction is paramount~\cite{kawaharazuka2025vision}. Therefore, there is an urgent need for an adaptor that preserves the generalization benefits of VLAs while incorporating the physical intelligence necessary for safe contact-rich manipulation.



In contrast, VIC provides compliant behavior in response to external forces through spring-damper dynamics, 
offers a complementary solution by modulating robot stiffness and damping to tackle interaction forces, enabling safe and adaptive physical interaction~\cite{ajoudani2018progress}. 
However, existing VIC approaches require manual tuning or task-specific parameter scheduling, lacking the semantic understanding to automatically adapt impedance based on visual context and task requirements.
Even advanced learning-based VIC methods~\cite{martin2019variable, 10517611} rely on extensive task-specific training or expert demonstrations, limiting their generalization to novel objects and scenarios.

Recent work has explored using VLMs to inform low-level control. For instance, SAS-prompt~\cite{amor2025sas} uses VLMs to provide in-context guidance for robot policies, while other studies have employed VLMs to generate control parameters~\cite{park2025making,jekel2025visio}. 
Additionally, VLMs have demonstrated their capability as in-context value learners~\cite{ma2024vision}, enhancing policy performance in various robotic tasks. 
Inspired by these advances, we extend this paradigm by employing VLM as an in-context impedance coach within a classical VIC, where stiffness and damping parameters are dynamically regulated through multimodal reasoning based on visual, contact force and linguistic context inputs~\cite{zhang2025safe}.

We present CompliantVLA-adaptor, a timely and effective but plug-and-play modular solution, before we train a high-performing VLA with sufficient and fine-grained datasets. 
The CompliantVLA-adaptor augments VLA models with VLM-guided variable impedance control for safe contact-rich manipulation. 
Our experimental findings demonstrate that CompliantVLA-adaptor not only enhances task success rates but fundamentally transforms the nature of failure modes. While baseline VLA models exhibit catastrophic failures characterized by contact force threshold violations and toppled objects, our approach achieves graceful degradation, with failures primarily due to minor misalignments or slippage rather than unsafe forces.
Specifically, we present the following contributions: 
\begin{itemize}
  \item The proposed CompliantVLA-adaptor leverages a VLM-enhanced VIC to endow VLA models with compliant capability to physical interaction, 
  where the VLM generates context-aware impedance parameters, translating high-level semantic understanding into low-level control parameters (stiffness, damping).  
  \item We develop a control system that seamlessly integrates VLM parameters generation with VIC execution, 
  maintaining the generalization benefits of VLM's reasoning while adding the safety improvement of compliant control. 
  \item We demonstrate our approach on several contact-rich tasks across simulation and real hardware,
    showing improvements over the state-of-the-art VLA baselines in both success rates and safety metrics.
\end{itemize}

\section{Related Work}

\subsection{Vision-language-action models for robotic manipulation}
Recent advancements in VLA models have significantly enhanced robotic manipulation capabilities by enabling robots to interpret and execute complex tasks described in natural language. 
Models such as RDT~\cite{liu2024rdt}, Pi0~\cite{black2024pi_0}, DiffusionVLA~\cite{wen2025diffusionvla}, Pi0.5~\cite{intelligence2025pi_}, OpenVLA~\cite{kim2024openvla}, and OpenVLA-oft~\cite{kim2025fine} leverage large-scale pretraining on diverse datasets to learn rich visual and linguistic representations, allowing them to generalize across a wide range of manipulation tasks. 
These models typically consist of a high-level policy that maps visual and linguistic inputs to desired action, which are then executed by a low-level controller.
For instance, OpenVLA employs an operational space controller (OSC) to apply the end-effector displacements generated by the VLA model.
InstructVLA~\cite{yang2025instructvla} further extends this paradigm by incorporating human feedback to refine action generation, but it still relies on a simple PD controller.
While these VLA models excel at semantic understanding and high-level action planning, they often lack consideration of the physical interaction dynamics involved in contact-rich tasks.
This limitation can lead to unsafe or failed interactions when robots encounter unexpected resistance or delicate objects, as they are unable to adapt their compliance based on the interaction forces.
\subsection{Variable impedance control for safe physical interaction}
Variable impedance control (VIC) enables robots to safely and adaptively perform contact-rich tasks by regulating compliance in response to interaction dynamics, inspired by human motor control \cite{yang2011human}. Unlike fixed-gain controllers, VIC improves robustness and safety in uncertain environments \cite{buchli2011learning} and is widely recognized as essential for physical interaction \cite{ajoudani2018progress}. 
Recent work further shows that using variable impedance actions in reinforcement learning (RL) enhances sample efficiency, robustness, and generalization \cite{martin2019variable}, and supports safe exploration in contact-rich scenarios \cite{10517611}. These studies establish VIC as a key paradigm for adaptable, safe, and efficient manipulation.

Learning-based VIC has advanced through RL approaches that incorporate force/torque feedback for online stiffness modulation \cite{martin2019variable,10517611,10401924,9833312}, imitation learning approaches that extract impedance profiles from human demonstrations or inverse RL \cite{wu2023impedance}, and context-aware methods leveraging human arm stiffness, trust models, or adaptive control \cite{9864178,LIAO2024102730}.
Despite these successes, existing VIC methods remain limited: RL demands task-specific training and risky exploration, imitation learning depends on expert demonstrations and struggles to generalize, and context-aware schemes rely on hand-crafted features that cannot exploit rich semantic cues (e.g., visually identifying fragile electronics).

Crucially, none of these can harness the broad semantic knowledge embedded in internet-scale vision-language models to inform impedance adaptation. This semantic gap forces conservative tuning or risks unsafe contact with novel objects. Our CompliantVLA framework closes this gap by bridging high-level semantic reasoning with low-level compliant control through VLM-guided variable impedance.


\subsection{VLMs enhance impedance parameter generation for safe physical interaction}
VLMs offer unprecedented capabilities for robotics by understanding scene semantics and task requirements from visual inputs, context injection and crucial knowledge to determine appropriate physical interaction parameters. 

Prior VIC parameter generation methods fall into three categories: 
(1) Rely on expert knowledge to pre-define impedance schedules \cite{11128409,11106445} for specific tasks, limiting adaptability to novel scenarios. 
(2) Learning-based methods like ImpedanceGPT \cite{batool2025impedancegpt} use language models for parameter selection but evaluate only on simplified pick-and-place tasks without complex contact dynamics. 
(3) Multimodal approaches incorporate additional sensors\cite{jekel2025visio} such as eye tracking for teleimpedance control, while \cite{bi2025vla} adds tactile feedback—but require specialized hardware that limits deployment scalability.

Our CompliantVLA-adaptor uniquely combines a VLM's semantic understanding with real-time force feedback to generate context-aware impedance parameters. 
Unlike previous works that use VLMs only for high-level planning or require task-specific training, 
we leverage pre-trained VLMs to directly map visual-linguistic context to impedance values, enabling zero-shot generalization across diverse contact-rich manipulation tasks while maintaining safety through force-regulated adaptation.

\section{Methodology}
\begin{figure}[!t]
    \centering
    \includegraphics[width=0.95\linewidth]{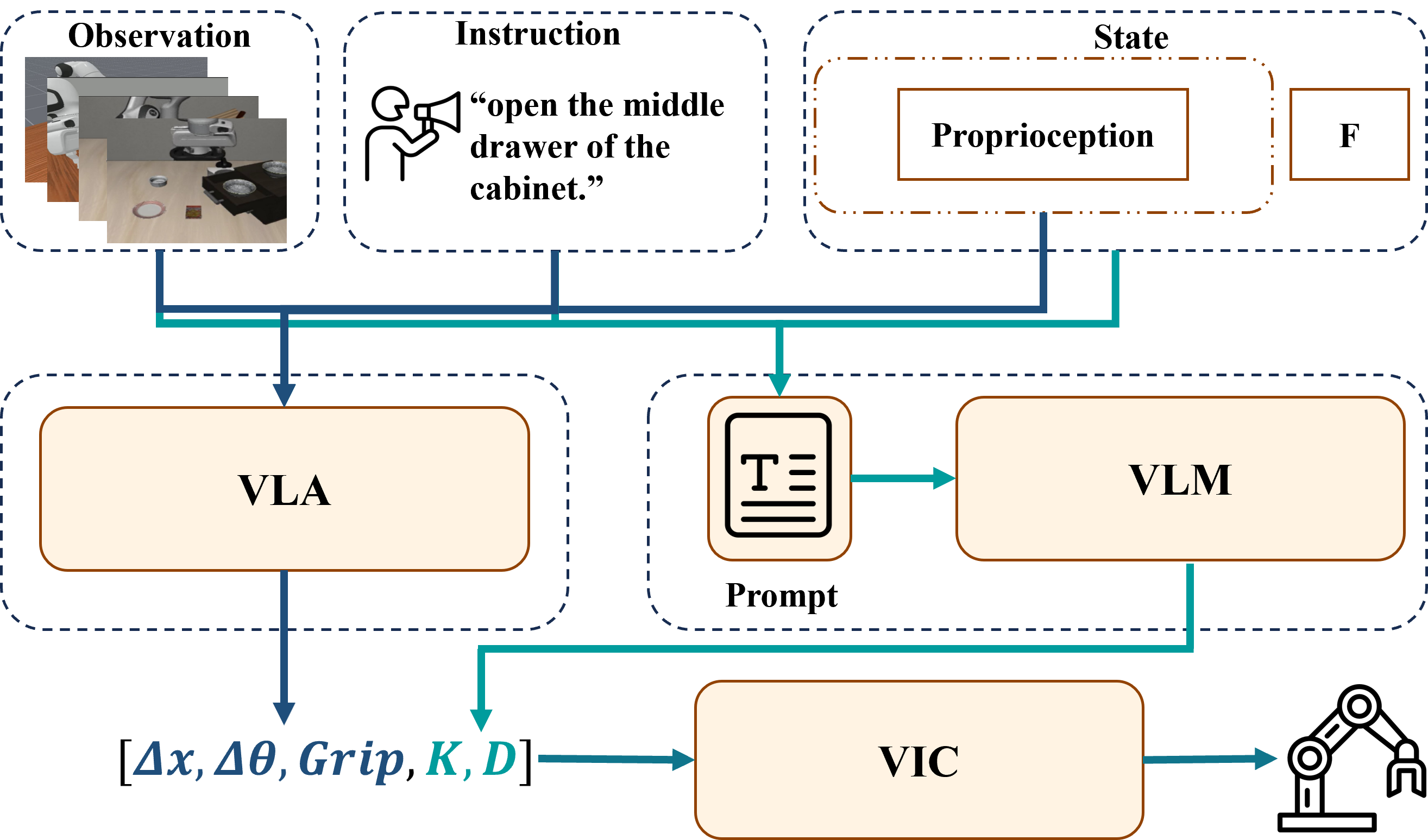}
    \caption{\small Overview of the CompliantVLA-adaptor. A VLM processes visual \textit{observations}, language instructions, and real-time force feedback~$\mathcal {F}$ to generate context-aware impedance parameters~$\mathcal{K, D}$. These parameters modulate a variable impedance controller(VIC) that executes actions generated by a VLA model, ensuring safe and adaptive contact-rich manipulation.}
    \label{fig:img_pipeline}
\end{figure}

We present CompliantVLA-adaptor, augmenting VLA models with VLM-enhanced context-aware variable impedance control. 
Our approach consists of three key components: (1) a hybrid VLA-VIC control architecture, (2) VLM-based impedance parameter generation from visual-linguistic context, and (3) a real-time force-regulated safety layer. Fig.~\ref{fig:img_pipeline} illustrates the complete system architecture.

\subsection{Problem formulation}

Consider a robotic manipulation task defined by the tuple $\mathcal{M} = (\mathcal{S}, \mathcal{A}, \mathcal{T}, \mathcal{C})$, where $\mathcal{S}$ represents the state space including visual observations, $\mathcal{A}$ denotes the action space, $\mathcal{T}$ is the task specification in natural language, and $\mathcal{C}$ represents physical constraints including force limits.

For a single 7 DoF robotic arm, the traditional VLA model maps observations and language instructions to actions, typically end-effector displacement values $\mathbf{x}_d \in \mathbb{R}^6$ and action for the gripper. However, this formulation ignores interaction forces $\mathbf{F} \in \mathbb{R}^6$, leading to unsafe contact interactions.

We augment an impedance-modulated control adaptor:
\begin{equation}
{CompliantVLA}: \text{VLM}(\mathcal{S} \times \mathcal{T} \times \mathcal{F}) \rightarrow (\mathcal{K}, \mathcal{D})
\end{equation}
 to VLA model:
\begin{equation}
 \pi_\text{VLA} : \mathcal{S} \times \mathcal{T} \rightarrow \mathcal{A}
\end{equation}
where a 6D vector $\mathcal{F} \in se^*(3)$ denotes the external contact force and torque, $\mathcal{K} \in \mathbb{R}^3$ and $\mathcal{D} \in \mathbb{R}^3$ represent the translational stiffness matrix $\mathcal{K}$ and the translational damping matrix $\mathcal{D}$, respectively. Where the translational stiffness $\mathcal{K}$ and translational damping $\mathcal{D}$ are denoted:
\begin{equation}
\mathcal{K} = \operatorname{diag}(k_x, k_y, k_z)
\end{equation}
\begin{equation}
\mathcal{D} = \operatorname{diag}(D_x, D_y, D_z)
\end{equation}
We set orientation stiffness matrices ${K}_o$, and orientation damping matrices ${D}_o$ with proportional coefficients shown in~\ref{eq:kodo}. This formulation enables the simultaneous generation of desired actions and compliance parameters.
\begin{equation} \label{eq:kodo}
K_o^i = \varepsilon k^i, \qquad 
D_o^i = 2 \zeta \sqrt{K_o^i}
\end{equation}
\noindent where $\varepsilon = 0.15$ and $\zeta = 0.707$. Here we set the proportional coefficients by empirical tuning to ensure the orientation stiffness and damping are appropriately scaled relative to the translational parameters, providing sufficient compliance for safe interaction while maintaining control stability.

The robot controller takes desired action $\mathcal{A}$, stiffness $\mathcal{K}$ and damping $\mathcal{D}$ as inputs to enhance the safety along the physical interaction.

\subsection{CompliantVLA-adaptor}
Our impedance generation module leverages a frozen VLM to extract multimodal features from external force, visual observations, and physics-injected language instructions, mapping them to context-aware impedance parameters.
To ensure safe interaction forces, we implement a dual-layer safety system that combines VLM-based parameter modulation with contact phase recognition. 

Specifically,
given an RGB image of wrist $\mathbf{I}_w$, image of full overview $\mathbf{I}_f$ and language instruction $\mathcal{T}$ and external force $\mathcal{F}$, (1) 
we query a pretrained VLM (e.g., ChatGPT-4o-mini) to reason about the current task execution phase \textit{current phase}; 
(2) integrate the suggested \textit{current phase} and prompt, VLM maps these features to impedance parameters.

\subsubsection{Contact Phase Recognition through multimodal information}
Vision-only phase detection presents inherent limitations, as VLMs cannot reliably detect contact states from visual information alone, particularly in scenarios with occlusions or subtle contact transitions. 
While force sensors provide direct contact information, relying solely on force feedback neglects valuable semantic context about expected interaction patterns. 
Therefore, we implement a hybrid approach that combines VLM-based visual understanding with force sensor feedback for robust phase detection.
With the following prompt(simplified, see details in our code), the VLM can distinguish between \texttt{Free-motion, Approaching, Contact, Retreat}. 
This semantic phase recognition eliminates the need for contact dynamics models while providing appropriate compliance for each interaction stage.

\begin{tcolorbox}[fontupper=\ttfamily\small, 
    title=Simplified Prompt for Contact Phase Recognition:]

You are a robotics expert capable of analyzing 
multimodal sensory inputs for recognizing the current task's contact phase.

Given \textcolor{red}{\{task description\}}, 
\textcolor{red}{\{force measurements\}}, 
\textcolor{red}{\{contact phase list\}}, 
determine the current task execution phase.

\textbf{Analyze the input to infer:} 
task execution context
\textbf{Output:} $phase = [phase]$

\end{tcolorbox}

\subsubsection{Multimodal-informed impedance parameter generation}


The final impedance parameters fed to the VIC controller combine VLM-generated values with real-time force feedback:
\begin{align}
\mathbf{K}_p^{\text{final}} &= \mathbf{K}_p^{\text{VLM}} \cdot \alpha_{\text{force}}\\
\mathbf{D}_p^{\text{final}} &= 2\sqrt{\mathbf{K}_p^{\text{final}} \cdot M_{\text{eff}}} \cdot \zeta
\end{align}
where $\alpha_{\text{force}} \in [0.2, 1]$ is a force-based scaling factor that reduces stiffness when measured forces exceed safe thresholds. The range of $\alpha_{\text{force}}$ is set by experience.
This ensures that even if the VLM suggests high stiffness based on visual context, the controller remains compliant in response to unexpected contact forces. $\mathbf{D}_p^{\text{final}}$ is computed to ensure critical damping, where $M_{\text{eff}}$ is the effective mass and $\zeta = 0.7$.

During execution, we continuously monitor force/torque sensor readings $\mathbf{F}_{\text{meas}}$ at 1000 Hz. 
The reactive regulation strategy adapts based on both measured forces and context reasoning from the VLM. 

Considering the complex contact and motion scenarios, 
we inject a physical-context prompt with our query prompts for impedance parameters 
so that the VLM can infer the appropriate anisotropic-centric impedance parameters.
The injected anisotropic-centric prompt is shown below:
\begin{tcolorbox}[fontupper=\ttfamily\small, 
    title=Simplified Prompt for context-aware impedance parameter generation:]

You are an expert robotic impedance controller capable of analyzing 
multimodal sensory inputs.

Given \textcolor{red}{\{task description\}}, current \textcolor{red}{\{phase\}} 
(Free\_motion, Approaching, Contact, or Retreat), 
\textcolor{red}{\{velocity\}} , and 
\textcolor{red}{\{force measurements\}}, 
\textcolor{red}{\{impedance range\}}, 
determine optimal anisotropic impedance parameters.

\textbf{Apply phase-based impedance hierarchy:}
\begin{itemize}
    \item \textbf{Free\_motion:} Highest impedance (precise position control)
    \item \textbf{Approaching:} Medium impedance (transitioning to compliance)
    \item \textbf{Contact:} Lowest impedance (maximum compliance)
    \item \textbf{Retreat:} Medium impedance (controlled withdrawal)
\end{itemize}

\textbf{Consider motion direction adaptation:}
\begin{itemize}
    \item \textbf{Primary motion axis:} Reduced impedance along intended motion direction
    \item \textbf{Constraint axes:} Elevated impedance perpendicular to motion for alignment maintenance
\end{itemize}

\textbf{Analyze the input to infer:} task requirement, current situation, 
and environmental physical constraints.

\textbf{Output:} $K = [K_x, K_y, K_z]$, $D = [D_x, D_y, D_z]$

where damping coefficients are proportional to stiffness values (10--20\%).

\end{tcolorbox}
This context-awareness prevents both unnecessary conservatism and dangerous over-forcing in primary motion direction.
Note that the impedance range should be appropriately scaled to match the hardware specifications.

\subsection{Hybrid VLA-VIC control architecture}

Our control architecture seamlessly integrates VLA action generation with VIC execution enhanced by VLM, 
maintaining the benefits of VLA generalization while adding compliant safety.

The system operates at three temporal scales:
\begin{itemize}
    \item {VLM-informed impedance generation ($\sim$ 1 Hz):} The VLM processes visual-linguistic context and force feedback to generate impedance parameters.
    \item {VLA action chunk ($\sim$ 3 Hz):} VLA generates desired end-effector displacements $\mathbf{x}_d$ based on current observations and instructions.
    \item {Low-level controller (1000 Hz):} VIC controller tracks desired poses with adaptive compliance to maintain safe contact.
\end{itemize}

\section{Experiments}
We evaluate CompliantVLA-adaptor on a suite of contact-rich manipulation tasks, such as object insertion and contact-sensitive pick-and-place. 
We compare our approach against the SOTA VLA models, which rely on position control without force adaptation. 

In these experiments, we aim to answer the following questions:
(1) Does CompliantVLA-adaptor improve task success rates compared to baseline VLA models under force thresholds?
(2) How effectively can the VLM generate context-aware impedance parameters for safe interaction?


\subsection{Simulation tasks }
All simulations are conducted in the same environment as the baselines to ensure fair comparison. 
To show the effectiveness of our CompliantVLA-adaptor, we compare our approach against three VLA models: Pi0 \cite{black2024pi_0}, RDT-1B \cite{liu2024rdt}, and OpenVLA-oft \cite{kim2025fine}.
Focusing on contact-rich tasks, we selected 8 representative tasks from the LIBERO \cite{liu2023libero} and ManiSkill benchmarks \cite{tao2024maniskill3}. All experiments were performed on 4 NVIDIA RTX A6000 GPUs.

\begin{figure}[!t]
    \centering
    \includegraphics[width=1\linewidth]{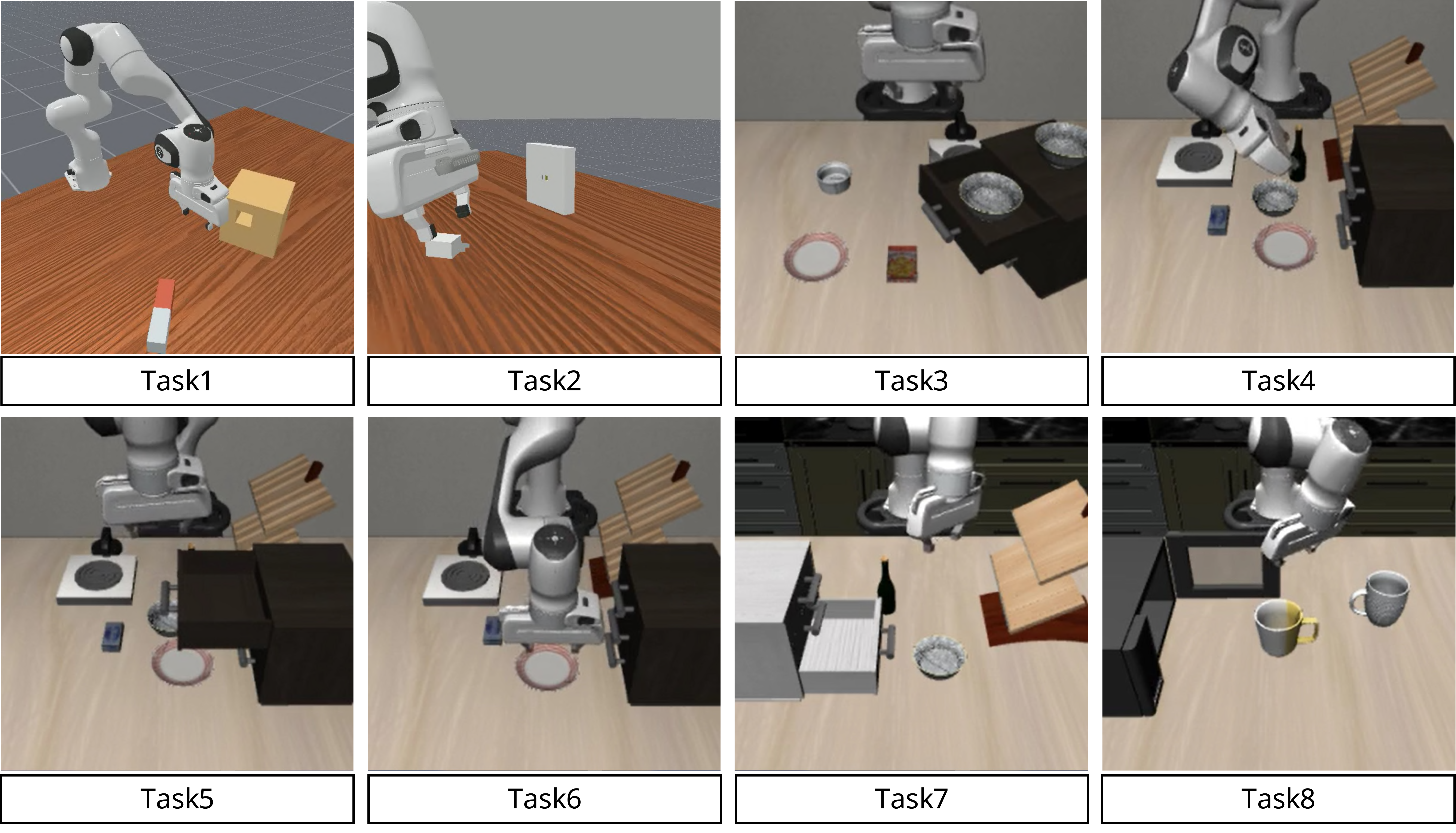}
    \caption{\small Contact-rich tasks in simulation. Task1 and Task2 are from ManiSkill for RDT, while the remaining tasks come from LIBERO for Pi0 and OpenVLA. Together, they form a contact-centric task suite.}
    \label{fig:tasks}
\end{figure}

\textbf{Task descriptions:} To comprehensively evaluate CompliantVLA-adaptor, we use eight representative contact-rich tasks from established benchmarks, instead of the full set of 100+ tasks, to avoid redundancy and resource constraints.
Table~\ref{tab:task_list_simulation} summarizes the task descriptions, including precision insertion, shape sorting, drawer opening/closing, and object placement in constrained environments shown in Fig.~\ref{fig:tasks}. 
These tasks are chosen to evaluate the CompliantVLA-adaptor's ability to handle complex contact dynamics that require both semantic understanding and safe physical interaction.

\textbf{Baseline models:} To diversely evaluate the effectiveness, we select three leading VLA models which are different architectures and training paradigms:
\textit{Pi0} \cite{black2024pi_0} uses flow matching for action generation;
\textit{RDT-1B} \cite{liu2024rdt}, the largest diffusion-based foundation model for robotic manipulation; 
and \textit{OpenVLA-oft} \cite{kim2025fine}, an Optimized Fine-Tuning (OFT) variant of OpenVLA.

\begin{table}
    \centering
    \caption{List of tasks in simulation}
    \begin{tabular}{cl}
    \hline
    \hline
        Task No. & Descriptions \\
        \hline
       Task 1 & (\textbf{Peg Insertion}) Pick up a orange-white peg and insert the \\
              & orange end into the box with a hole in it\\
       Task 2 & (\textbf{Plug Charger}) Pick up one of the misplaced shapes on the \\
              & board/kit and insert it into the correct empty slot\\
       Task 3 & (\textbf{Drawer Env}) pick up the black bowl in the top drawer of \\
              & the wooden cabinet and place it on the plate\\
       Task 4 & (\textbf{Open Drawer}) open the middle drawer of the cabinet\\
       Task 5 & (\textbf{Open Drawer}) open the top drawer and put the bowl inside\\
       Task 6 & (\textbf{Push Plate}) push the plate to the front of the stove\\
       Task 7 & (\textbf{Close Drawer}) put the black bowl in the bottom drawer of \\
          & the cabinet and close it\\
       Task 8 & (\textbf{Microwave Env}) put the yellow and white mug in the\\ 
       & microwave and close it\\
        \hline
    \end{tabular}
    
    \label{tab:task_list_simulation}
\end{table}

\textbf{Experiment protocol:} We adopt a two-stage evaluation protocol to isolate the contribution of CompliantVLA-adaptor under the same contact force safe constraint for each stage, see Sec.~\ref{sec:contact_force_measurement}:
(1), performance of baseline VLA model and (2) augmented operation with our VLM-enhanced VIC controller.

\textit{Stage 1 - Baseline VLA Performance:}  We evaluate each baseline VLA model using its default control strategy. 
Since all three models employ position- or trajectory-based control without explicit force feedback, there is no compliance adaptation during contact.
Therefore, we set a strict force threshold of 30N to ensure safety during execution. A task is terminated after three consecutive threshold violations or if not completed within the time limit (see Sec.~\ref{sec:contact_force_measurement}).

\textit{Stage 2 - CompliantVLA-adaptor Integration:} We augment each baseline VLA model with the proposed CompliantVLA-adaptor,
replacing its low-level controllers with our VLM-enhanced VIC controller
for compliant execution. In the meantime, we keep the same force check criteria for fair comparison (see Sec.~\ref{sec:contact_force_measurement}).


\subsection{Contact force measurement and safety criteria} \label{sec:contact_force_measurement}
Contact forces are measured using a simulated wrist-mounted force/torque sensor. The sensor provides real-time feedback on interaction forces during task execution.
Safety criteria are defined based on task requirements, with force thresholds set to prevent excessive contact that could damage objects or the robot. For all selected contact-sensitive tasks, the force threshold is set to 30N.

During each task execution, we monitor force violations across all dimensions (
$F_x, F_y, F_z$
). The critical 30N threshold is based on typical safe interaction forces. 
If the contact force exceeds this threshold three times consecutively (to filter out transient spikes due to sensor noise)
during a single trial, the task is immediately terminated and marked as failed to prevent potential damage. 

\subsection{Contact force regulation by CompliantVLA-adaptor}
We further evaluate the effectiveness of our VLM-based impedance parameter generation for regulating contact forces in contact-rich manipulation tasks.
Although force is continuously monitored during the whole task execution, impedance parameters are updated by querying the VLM every two action steps (${\approx}$0.5s), balancing responsiveness with computational load.



\subsection{Real-world experiments}
\begin{figure}[tbh]
    \centering
    \includegraphics[width=0.9\linewidth]{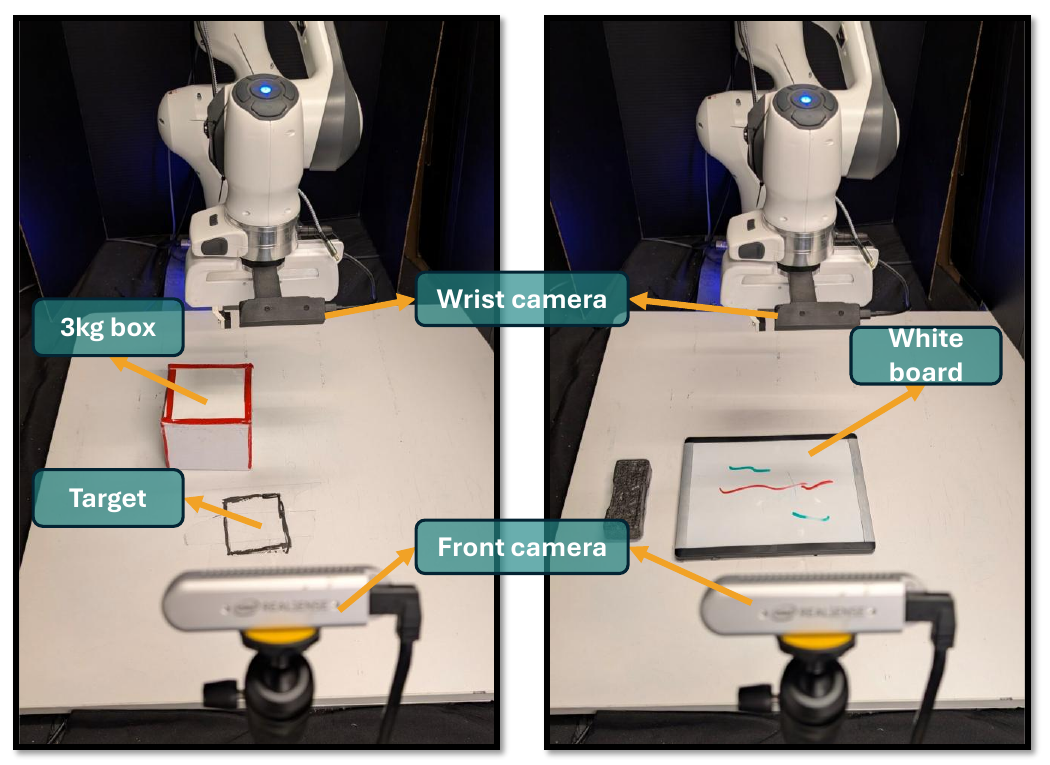}
    \caption{\small Real-world experiment setups.  Left: box-pushing setup, where the robot pushes a 3\,kg box toward a marked target region. Right: whiteboard-wiping setup, where the robot removes marker traces on the board.}
    \vspace{-1em}
    \label{fig:real-setup}
\end{figure}

Our experiment setup consists of a 7-DoF Franka Emika Panda robot arm with a Franka Panda parallel gripper. The same hardware is used in all real-world experiments to ensure consistency.
The robot operates in a workspace with diverse objects for contact-rich manipulation tasks (see Fig.~\ref{fig:real-setup}). We completed the ``\textit{push the heavy box to the target position}"  and ``\textit{wipe the whiteboard and clean the mark}".

To facilitate efficient visual perception, real-time control and model inference, we designed and implemented a dual-machine system: the client (Ubuntu 22.04 with real-time kernel) handles data acquisition and low-level control, while the server (NVIDIA RTX A6000, 48GB VRAM) hosts the VLA model.



\subsubsection{Visual perception}
For visual perception hardware, we employed a dual-camera configuration to capture comprehensive scene information. A Logitech USB camera serves as the main camera, providing a global view of the robot arm and the workspace. 
An Intel RealSense D435i depth camera functions as the wrist camera, mounted on the end-effector of the Franka Emika Panda arm using an open-source, 3D-printed fixture, 
offering an ego-centric view. To meet the VLA's input requirements, images from both cameras are uniformly resized to $224\times224$ pixels before being fed into the model. 
\subsubsection{Data collection for fine tuning} We also collect real-world data for fine-tuning the VLA model, which includes RGB images from both cameras, proprioceptive states, and language instruction during task execution.
50 successful trials are collected for each task, which will be used for fine-tuning the Pi0.5 VLA backbone to enhance its performance in real-world scenarios.



\begin{figure*}[tbh]
    \centering
    \includegraphics[trim=0.1cm 3cm 0.1cm 3cm,width=0.97\linewidth]{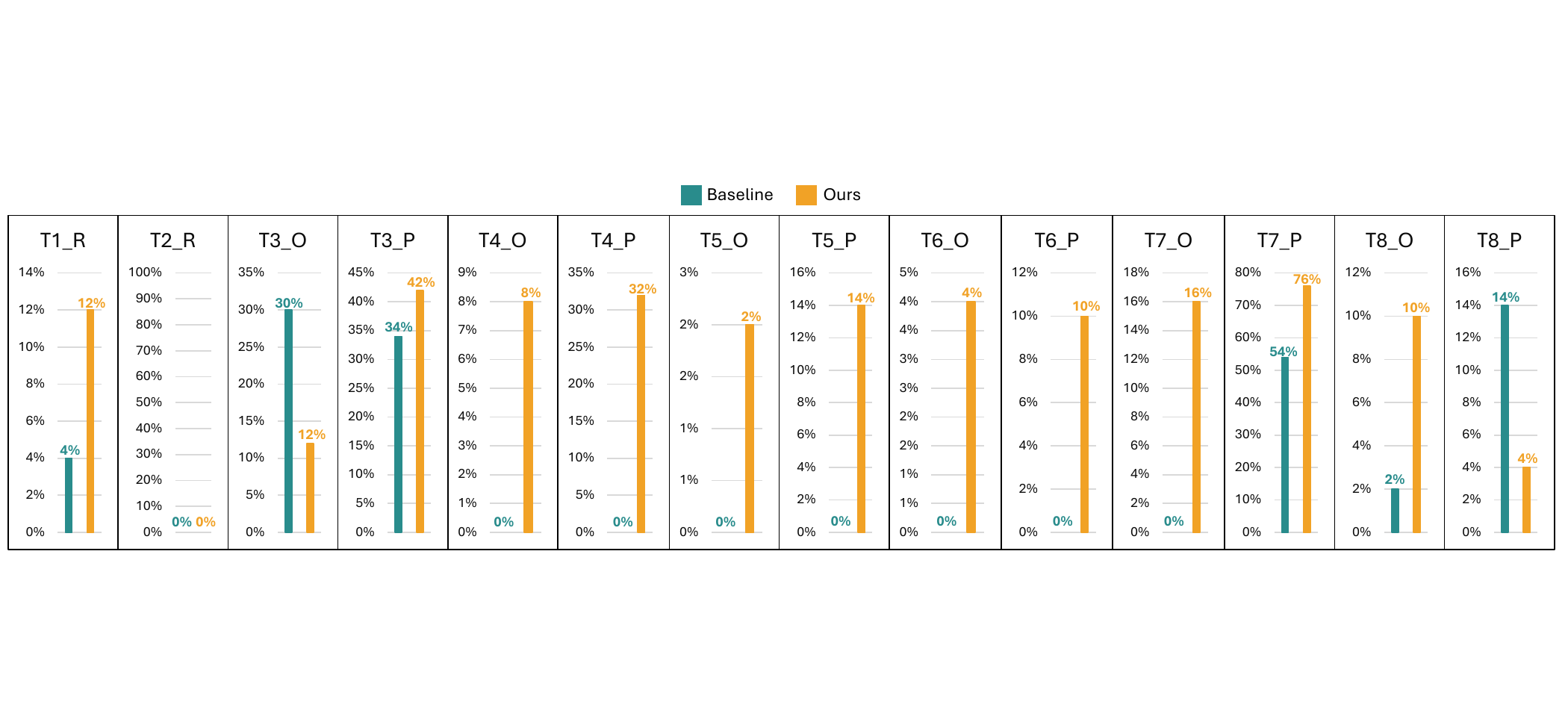}
    \vspace{-1em}
    \caption{\small Evaluation results of task success rates under contact force constraint across 8 related tasks in simulation.
    Title of each bar diagram indicates a different setup, where ``T" with a value denotes task number, see its description in Tab.~\ref{tab:task_list_simulation}, 
    and the suffix ``-R", ``-P" and ``-O" means using RDT, Pi0 and OpenVLA-oft model, respectively.
    Different colors indicate the run with and without our CompliantVLA-adaptor.
    }
    \label{fig:compliant_vla_accuracy_fig}
\end{figure*}

\section{Results}
Through comprehensive simulation and real-world experiments, we demonstrate the effectiveness of the proposed CompliantVLA-adaptor that enhances the safety in contact-rich tasks compared to state-of-the-art VLA baselines,
especially in scenarios requiring fine force modulation and compliance adaptation.

\subsection{Simulation results}
We present quantitative results from our simulation tasks shown in Fig.~\ref{fig:compliant_vla_accuracy_fig}, comparing the performance of our CompliantVLA-adaptor with the SOTA VLA baselines. 
The results show that our approach achieves significantly higher task success rates and reduces force violations across most of the contact-rich manipulation tasks (7/8). 
The adaptive impedance parameters enable the robot to maintain safe interactions while effectively completing the tasks.

Fig.~\ref{fig:compliant_vla_accuracy_fig} presents a comprehensive comparison between the baseline VLA models and our CompliantVLA-adaptor approach.
The title of each bar diagram indicates different setups, where ``T" with a value denotes task number (see its description in Tab.~\ref{tab:task_list_simulation}), 
and the suffix ``-R", ``-P" and ``-O" means using RDT~\cite{liu2024rdt}, Pi0~\cite{black2024pi_0} and OpenVLA-oft~\cite{kim2025fine} model, respectively.
Different colors indicate the run with or without our CompliantVLA-adaptor.
The y-axis indicates the task success rates under the contact force threshold of 30N.
The results show that the baseline VLA models exhibit highly unstable performance across the task suite, even worse 0\% in some tasks.
Our CompliantVLA-adaptor improves performance across most tasks, demonstrating its effectiveness.

Note that the baseline VLA models exhibit highly unstable performance across the task suite, with a maximum success rate of only 54\% and some tasks always failed if considering contact force during execution.
This shows the different performance if considering the force threshold or not, compared to the original papers.

After integrating the CompliantVLA-adaptor into these VLA models, most of the evaluated tasks showed consistent improvements while staying safe, with a maximum success rate of only 76\% and significant improvement among tasks that always failed in baselines.
The average success rate across all tasks increases from 9.86\% to 17.29\%, 
this aggregate improvement, while substantial, varies significantly across task categories. 
Tasks involving mechanical constraints (drawers, stove knobs) show the most dramatic relative improvements, as these scenarios particularly benefit from the adaptive compliance provided by our VLM-guided impedance control. 
Conversely, tasks with higher baseline performance, such as simple pick-and-place operations, show more modest but still meaningful improvements, primarily through enhanced force safety during object grasping and release phases.
These results validate our hypothesis that augmenting state-of-the-art VLA models with semantic-aware compliant control can bridge the gap between high-level task understanding and low-level force-safe execution.

In terms of failure cases, baseline VLA models frequently encounter force violations leading to task termination, particularly in \cite{liu2024rdt} plug charger task (Task2 in Tab.~\ref{tab:task_list_simulation}) 
and both \cite{black2024pi_0} and \cite{kim2025fine} in drawer manipulation tasks (Task3 ~ 5 in Tab.~\ref{tab:task_list_simulation}) and \cite{kim2025fine} in push task(Task6 in Tab.~\ref{tab:task_list_simulation}).
In contrast, our CompliantVLA-adaptor significantly reduces these violations by dynamically adjusting impedance parameters based on real-time force feedback and task context.
This adaptive behavior allows the robot to navigate contact-rich environments more safely and effectively, ultimately leading to higher task success rates.

\subsection{Real-world experiment results}

\begin{figure}[h]
\centering
\begin{subfigure}{0.98\linewidth}
        \centering
        \includegraphics[trim=0cm 0.1cm 0cm 0.1cm,width=1\linewidth]{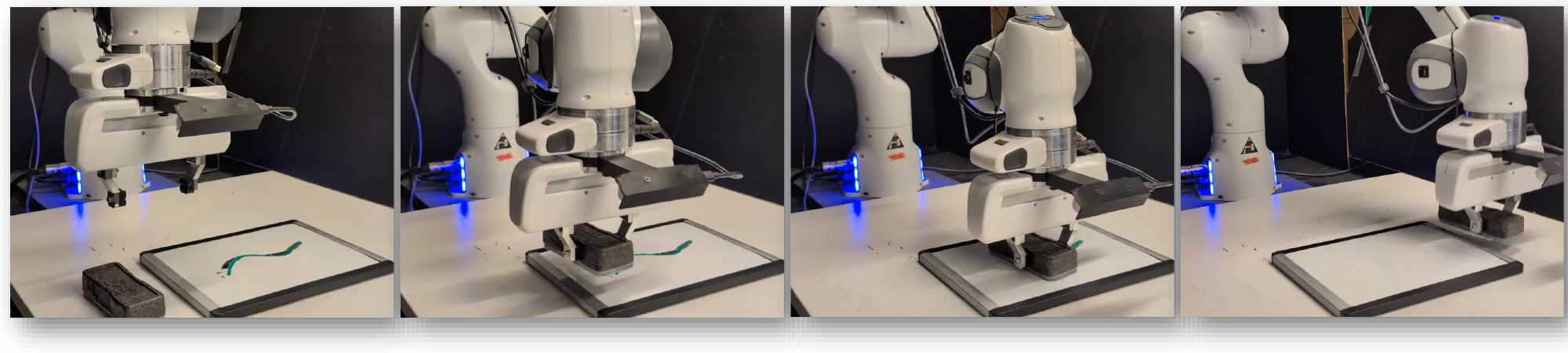}
    \end{subfigure}
    \begin{subfigure}{0.99\linewidth}
        \centering
        \includegraphics[trim=0cm 0.1cm 0cm 0.1cm,width=1\linewidth]{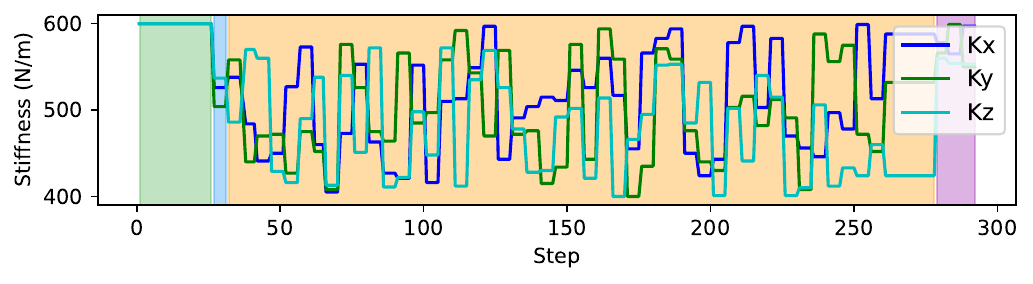}
    \end{subfigure}
    \begin{subfigure}{0.99\linewidth}
        \centering
        \includegraphics[trim=0cm 0.1cm 0cm 0.1cm,width=1\linewidth]{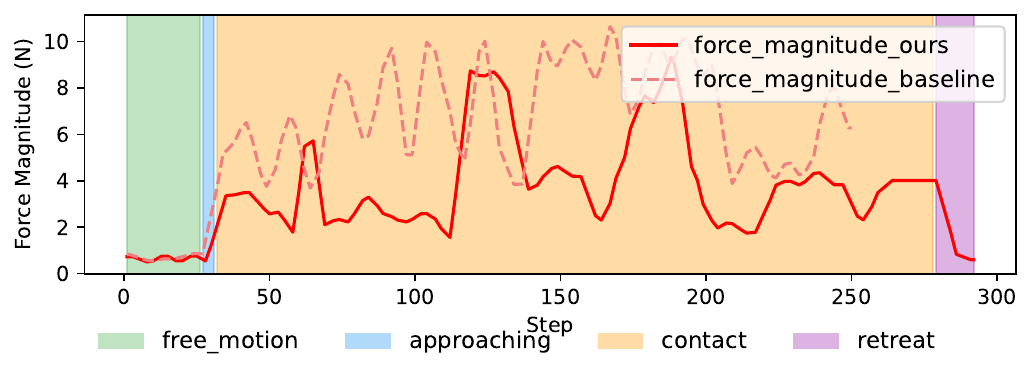}
    \end{subfigure}
    \begin{subfigure}{0.98\linewidth}
        \centering
        \includegraphics[trim=0cm 0.1cm 0cm 0.1cm,width=1\linewidth]{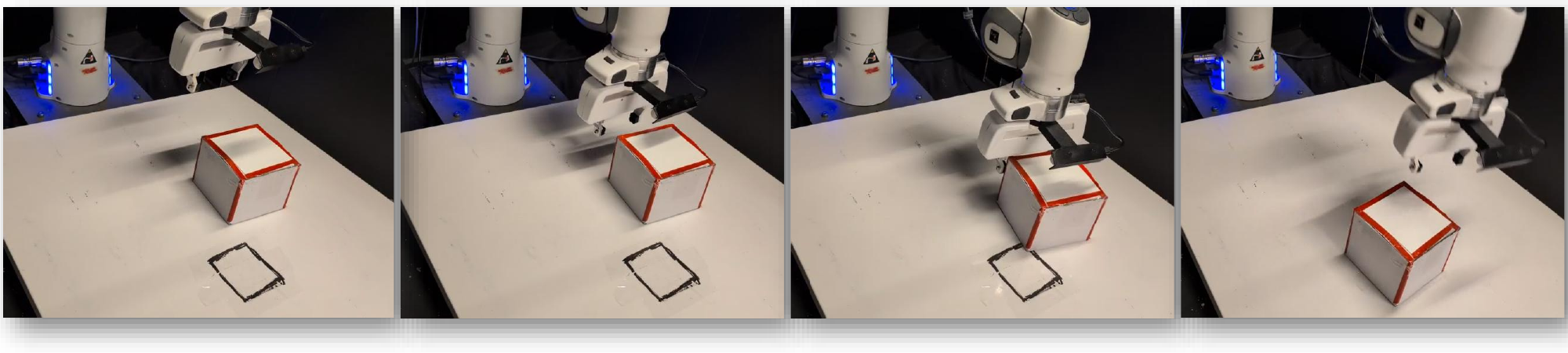}
    \end{subfigure}
    \begin{subfigure}{0.99\linewidth}
        \centering
        \includegraphics[trim=0cm 0.1cm 0cm 0.1cm,width=1\linewidth]{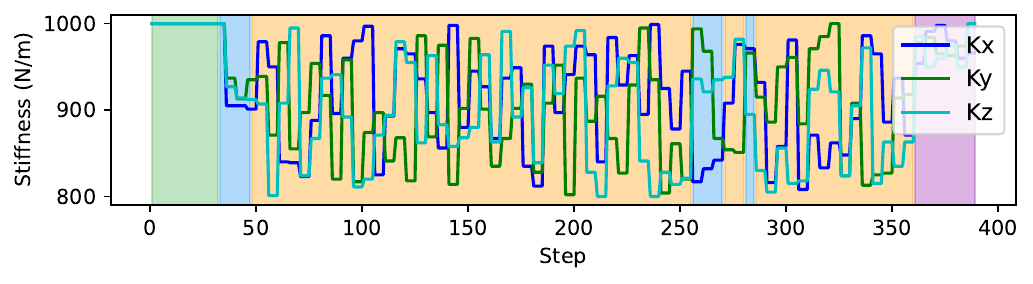}
    \end{subfigure}
    \begin{subfigure}{0.99\linewidth}
        \centering
        \includegraphics[trim=0cm 0.1cm 0cm 0.1cm,width=1\linewidth]{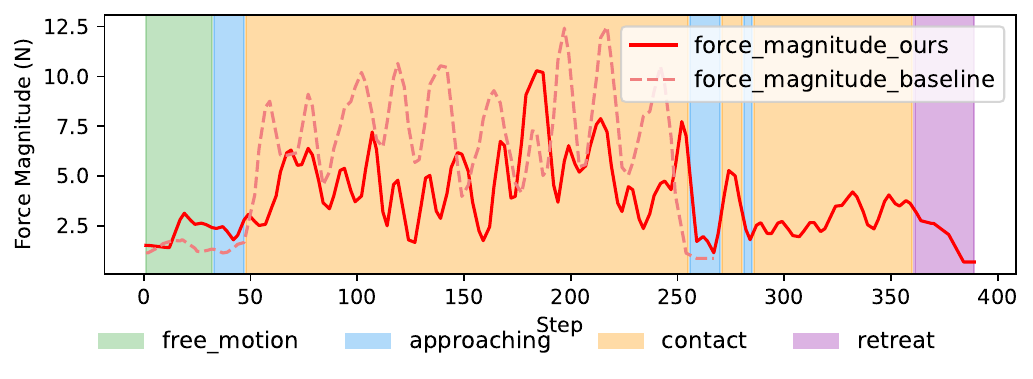}
    \end{subfigure}

  \caption{\small Real-world experiment evaluation: stiffness regulation and measured force during the task execution. The upper 4 sub-figures in a row show the task process, while the curves below each task illustrate the adaptive impedance control and contact force during the task, with background colors indicating distinct task phases.}
    \label{fig:real-exp}
\end{figure}

As shown in Fig.~\ref{fig:real-exp}, the adaptor consistently avoids unsafe forces, demonstrating its effectiveness as a safety module. The stiffness and force profiles indicate that the robot maintains compliant interactions, adjusting its impedance parameters in response to contact conditions. The interaction force remains within safe limits throughout the task execution, validating the safety benefits of our approach in real-world scenarios.
In the wiping task, the robot successfully modulates its stiffness in range of $[400,600]$ to maintain gentle contact with the whiteboard, while in the pushing task, it adapts its impedance to safely push the box with higher stiffness in $[800,1000]$. 
We also observe that the force profiles show smooth transitions without abrupt spikes, indicating that the robot is effectively regulating contact forces to prevent damage to the environment and itself.
    

\section{Discussion and Limitations }
We discuss the limitations of our current approach and potential avenues for future work.
While our CompliantVLA-adaptor demonstrates significant improvements in safe contact-rich manipulation, several limitations remain: 
\begin{enumerate*}
    \item VLM inference is slower than the servo loop, it is not on the safety-critical path: abnormal-contact handling remains in the 1000 Hz force-regulated VIC, while VLM latency only limits the refresh rate of the semantic impedance prior.
    \item using the API calls is expensive and not eco-friendly. On the other hand, using onboard VLM is computationally heavy.
    \item VLMs may not generalize well to unseen objects or environments, leading to suboptimal performance in novel situations.
\end{enumerate*}

\section{Conclusion}
In this work, we propose a CompliantVLA-adaptor that leverages physical context-informed VLMs with VIC to enhance the performance of VLA in safe and effective manipulation in contact-rich tasks. 
By leveraging the semantic grounding of VLMs and the physical robustness of VIC, we enable robots to adapt their impedance parameters based on task context derived from visual and language inputs. 
Our experimental results validate the effectiveness of the proposed approach in both simulation and real-world experiments, achieving higher task success rates and reducing force violations. 
 This adaptor provides a lightweight, plug-and-play safety layer for existing VLA models, offering a practical step toward safe and generalizable contact-rich manipulation, bridging high-level semantic reasoning and low-level compliant control, enabling safer manipulation for VLA models in contact-rich environments.

\bibliographystyle{IEEEtran}
\bibliography{main}
\end{document}